\documentclass{article} 
\usepackage{colm2024_conference}

\usepackage{microtype}
\usepackage{hyperref}
\usepackage{url}
\usepackage{booktabs}
\usepackage{rotating}
\usepackage{multirow}
\definecolor{darkblue}{rgb}{0, 0, 0.5}
\hypersetup{colorlinks=true, citecolor=darkblue, linkcolor=darkblue, urlcolor=darkblue}
\usepackage{graphicx}
\usepackage{wrapfig}

\definecolor{darkblue}{rgb}{0.0, 0.0, 0.55}

\definecolor{gray}{HTML}{708090}
\definecolor{gem}{HTML}{80A8EE}
\definecolor{claude}{HTML}{cc785c}
\definecolor{darkgreen}{HTML}{0FA37F}
\definecolor{lightgreen}{HTML}{51DA4C}
\definecolor{mistral}{HTML}{FE4A00}
\definecolor{llama}{HTML}{0068E8}

\title{Can Large Language Models do Analytical Reasoning?}

\author{Yebowen Hu$^\star$, Kaiqiang Song$^\dag$, Sangwoo Cho$^\dag$, Xiaoyang Wang$^\dag$\\
\textbf{Hassan Foroosh$^\star$, Dong Yu$^\dag$, Fei Liu$^\diamond$}\\[0.5em]
$^\star$University of Central Florida \, 
$^\dag$Tencent AI Lab, Bellevue, WA \, 
$^\diamond$Emory University\\[0.5em]
\texttt{\{yebowen.hu, hassan.foroosh\}@ucf.edu}\\
\texttt{\{riversong, swcho, shawnxywang, dyu\}@global.tencent.com }\\
\texttt{fei.liu@emory.edu}
}

\colmfinalcopy 
\begin{document}

\maketitle

\begin{abstract}
This paper explores the cutting-edge Large Language Model with analytical reasoning on sports.
Our analytical reasoning embodies the tasks of letting large language models count how many points each team scores in a quarter in the NBA and NFL games.
Our major discoveries are in two folds.
Firstly, we find among all the models we employed, GPT-4 stands out in effectiveness, followed by Claude-2.1, with GPT-3.5, Gemini-Pro, and Llama-2-70b lagging behind.
Specifically, we compare three different prompting techniques and a divide-and-conquer approach, we find that the latter was the most effective.
Our divide-and-conquer approach breaks down play-by-play data into smaller, more manageable segments, solves each piece individually, and then aggregates them together.
Besides the divide-and-conquer approach, we also explore the Chain of Thought (CoT) strategy, which markedly improves outcomes for certain models, notably GPT-4 and Claude-2.1, with their accuracy rates increasing significantly.
However, the CoT strategy has negligible or even detrimental effects on the performance of other models like GPT-3.5 and Gemini-Pro.
Secondly, to our surprise, we observe that most models, including GPT-4, struggle to accurately count the total scores for NBA quarters despite showing strong performance in counting NFL quarter scores.
This leads us to further investigate the factors that impact the complexity of analytical reasoning tasks with extensive experiments, through which we conclude that task complexity depends on the length of context, the information density, and the presence of related information.
Our research provides valuable insights into the complexity of analytical reasoning tasks and potential directions for developing future large language models.
\end{abstract}
\section{Introduction}
\label{sec:intro}


In recent years, advancements in large language models have been remarkable, with OpenAI's ChatGPT~\citep{OpenAI2023} leading the way in demonstrating exceptional capabilities across a broad spectrum of natural language processing tasks.
These tasks encompass information extraction~\citep{lu2023pivoine,liu2023mmc}, question answering~\citep{tonmoy2024comprehensive, tan2023chatgpt}, natural language understanding~\citep{mao2023gptdriver,LIU2023}, text generation~\citep{sun2024trustllm,10.1145/3591196.3596612}, translation~\citep{fan2020englishcentric, wu2024adapting}, and summarization~\citep{liu2023learning, hu-etal-2023-meetingbank}. 
Additionally, ChatGPT exhibits a profound capacity to manipulate other data formats, including tables and code.
Competing models, such as Anthropic's Claude~\citep{claude-2.1}, Google's Gemini~\citep{geminiteam2023gemini}, Meta's Llama~\citep{touvron2023llama2}, and Mistral's Mixtral~\citep{mixtral-8x7b} have also displayed performance levels that are on par with benchmarks, showcasing their significant contributions to the field.
However, such models remain imperfect as they lack some tasks like math~\citep{zhu-etal-2023-solving, cobbe2021training, hendrycks2021measuring} and complex reasoning~\citep{masry2022chartqa, chen2022finqa}.
\begin{figure}[htbp]
    \centering
    \includegraphics[width=0.8\linewidth]{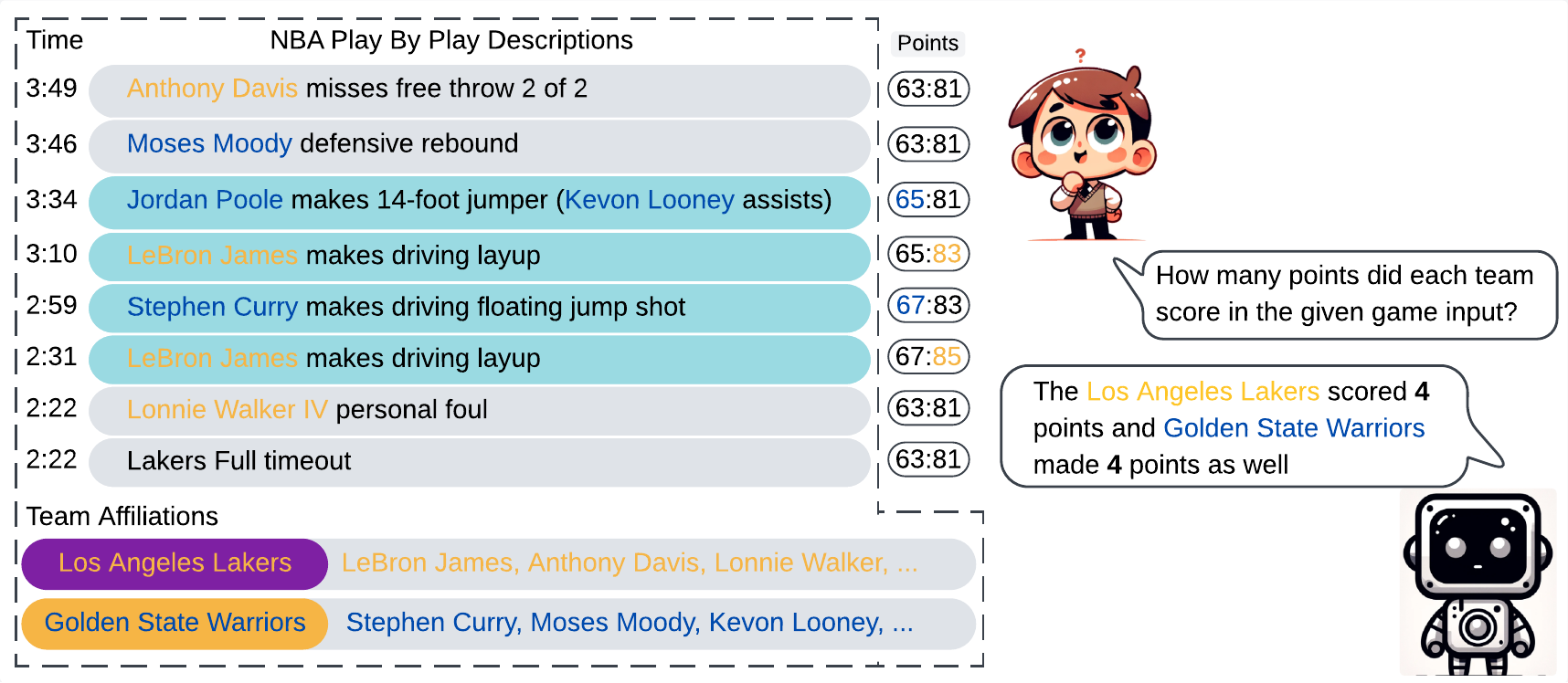}
    \caption{\small Play-by-play descriptions of an NBA game, including timestamps, play-by-play descriptions, team affiliations and total points. The content indicated in dot circles are inputs for our task. The total points for both teams are withheld from LLMs. Highlighted descriptions are scoring moves originally labeled in source data. This image was created with the assistance of DALL·E}
    \label{fig:example-data}
    \vspace{0.2in}
\end{figure}
To evaluate the capabilities of recent cutting-edge language models in interpreting and analyzing sports statistics, our research focused on examining play-by-play data from the NBA and NFL, as provided by ESPN.
The primary goal was to assess the models' ability to answer straightforward analytical questions, such as determining the total points scored by each team within a specific quarter.
Figure~\ref{fig:example-data} illustrates an example of the task at hand, where both the team affiliations and the detailed play-by-play descriptions are provided.
The AI system is then asked to count the total points scored by both teams in the game clip.

Upon analyzing NBA data, it was observed that the majority of the models were unable to accurately respond to the queries posed.
Our findings also revealed a significant disparity in performance among the evaluated models.
Remarkably, GPT-4 emerged as the only model capable of delivering a somewhat accurate response to the posed questions, achieving an accuracy rate of 11\%.
In contrast, other models, including Claude-2.1, GPT-3.5, Gemini-Pro, and Llama-2-70b-chat, were found to have accuracy rates of less than 5\%.
Furthermore, we observed that incorporating a "chain of thought" approach markedly improved outcomes for certain models.
Specifically, GPT-4 and the Claude model benefited substantially from this technique, with their accuracy rates increasing to 40\% and 17\%, respectively.
Conversely, this strategy yielded negligible or even detrimental effects on the performance of the other models.

Additionally, rather than solely relying on the models, we sought to incorporate human insights through a divide-and-conquer method.
This approach involves a program to segment play-by-play data into smaller, manageable segments.
Subsequently, language models serve as executors to process these segments, extracting partial results.
These partial results are then aggregated to formulate the final results.
Our findings indicate a notable enhancement in performance through this approach, however, it continues to exhibit some limitations when applied to NBA games.

Furthermore, analyses of NFL quarter results reveal consistent patterns across models and methodologies.
Yet, as the complexity of the task decreases, the benefits of sophisticated methods diminish, rendering them unnecessary.
These observations amplify our intent of understanding the factors that contribute to the difficulty of analytical reasoning tasks and understand why models struggle with these challenges.

To achieve this goal, we formulated several hypotheses and carefully tested them through extensive experiments, focusing on a range of factors.
Our detailed analyses indicate that context length and information density significantly influence task complexity. 
Moreover, it has been noticed that including relevant information, even if not essential, improves the model's effectiveness.

To conclude, we summarize our contributions into three parts:
\begin{itemize}
    \item We conducted experiments on NBA and NFL play-by-play data to investigate the analytical reasoning capabilities of LLMs.
    According to empirical evidence, the current state-of-the-art language models (LLMs) are not capable of effectively solving such tasks.
    \item Through extensive experiments, we found the context length, the information density and the presence of related information play an important role in affecting the task difficulty.
    \item We propose that analytical reasoning can be viewed as a common assessment task for evaluating the planning abilities of LLMs. 
    Our analysis of the divide-and-conquer method shows that the skill of step planning is crucial for improved performance.
\end{itemize}

\section{Related Work}
\label{sec:related}

\textbf{Analytical Reasoning}: Previous works on mathematical reasoning of LLMs emphasis on solving math world problems. Datasets include SVAMP \citep{patel2021nlp}, GSM8k \citep{cobbe2021training} and PRM800K \citep{lightman2023lets} aggregate problems ranging from elementary to graduate level, benchmark LLMs according to reasoning paths and numerical results. Our proposed analytical reasoning task is distinguished from those works in terms of question length and reasoning complexity. We expect LLMs to perform contextual understand on sports descriptions and consistent addition operations across lengthy game input. 
\textbf{Prompting Methods}: As LLMs scale up, natural language instructions have become the dominant method to steer arithmetical ability of model \citep{wei2022finetuned}. Chain-of-Thought(CoT) \citep{wei2023chainofthought} first presents a significant improvement on numerical reasoning tasks through step-by-step reasoning process. Step-back prompting \citep{zheng2023step} illustrate an abstraction principle step before reasoning. SKiC \citep{chen2023skillsincontext} introduce a prompting framework to guide LLMs to compose basic skills to solve complex problem. Our work focuses on investigating whether LLMs are capable of handling analytical reasoning instead of prompting optimization. Therefore, we conducted experiments utilizing natural language instructions combined with formatting constraints or CoT for comparison. 
\textbf{Sports Analysis}: Previous works using sports data for news and live commentary generation\citep{wang2021knowledge, wang2022goal, huang-etal-2020-generating}. 
Sports-QA propose a dataset for bridging the gap between multi-modality resources in sports analysis \citep{li2024sportsqa}. 
\cite{hu2024sportsmetrics} aggregate a benchmark, SportsMetrics, as test bed to evaluate LLMs
long context information fusion capability. In our work, we introduce a novel reasoning task, sports analytical reasoning, and provide a detail analysis on possible mechanisms underlying the reasoning process of LLMs.
\section{Methodology}
\label{sec: method}

\subsection{Data \& Task}
To evaluate the analytical reasoning abilities of advanced language models, we conducted a study utilizing sports clips and corresponding play-by-play data sourced from ESPN.
These play-by-play data was well-crafted by ESPN's sports journalists to describe temporal event that occurred on court. In this work, we employed \textbf{SportsMetrics} dataset, which was original aggregated by \cite{hu2024sportsmetrics}.


We randomly sampled 100 NBA games out of a total of 28,492 collected matches and 100 NFL games out of 5,867. 
As depicted in Figure~\ref{fig:example-data}: the data includes a ``\emph{time}`` column indicating the specific moment of each play within the game, and a ``\emph{play}`` column detailing the actions occurring at those times.
Scoring plays that affect the game's score are highlighted by ESPN, though these particular labels are not disclosed to the language models in our experiments.
Team affiliation is also accessible from dataset.


\begin{table}[htbp]
    \centering
    \small
    \begin{tabular}{lccccc}
    \toprule[1pt]
     & \# Plays & \# Scoring Plays & \# Tokens & \# Athletes\\
    \midrule
    NBA & 116.47 & 28.78 & 1,557 & 17.69 \\
    NFL & 42.68 & 1.98 & 1,459 & 39.60 \\
    \bottomrule[1pt]
    \end{tabular}
    \caption{\small \textbf{Quarter-level} statistics are analyzed based on 100 games selected from each sport.}
    \label{tab: data statistics}
\end{table}

In our study, we segmented each game into quarters based on labels provided by ESPN.
As detailed in Table~\ref{tab: data statistics}, we present test set statistics in both NBA and NFL games.
The high number of short play descriptions and scoring plays in NBA games can be attributed to the fast-paced, continuous nature of basketball.
In contrast, NFL games, with longer play description and fewer scoring plays per quarter, reflect the more strategic, play-by-play nature of American football.
This may indicate the fundamentally natural different between NBA and NFL. 

In this work, we will focus on ``\textsc{points}`` as our main experiments, representing the scores accrued by both teams at the end of given game input.
Reasoning on ``\textsc{points}`` requires model to understand contextual meaning and perform consistent math reasoning ability. So we regard it as a sufficient task to illustrate analytical capability of LLM. Results analysis will be discussed in later section.


\subsection{Applied Models}
\label{ssec: models}
To better evaluate the performance on sports analytical reasoning tasks mentioned above, we employ five of the most cutting-edge conversational AI systems containing both open and closed-sourced models.
These include OpenAI's GPT-3.5-turbo-1106~\citep{ouyang2022training}, GPT-4-1106-preview~\citep{gpt4-turbo}:, Anthropic's Claude-2.1~\citep{claude-2.1}, Google's Gemini-Pro~\citep{geminiteam2023gemini} and Meta's Llama-2-70b-chat~\citep{touvron2023llama2}.
The cost to call GPT API is based on the pricing plan provided by OpenAI. GPT-4-1106-preview costs \$0.01 for input and \$0.03 for completion per 1,000 tokens. GPT-3.5 is more cost-effective which costs \$0.001 for input and \$0.002 for completion. The overall expenditure for 100 games on a single task cost around \textbf{\$12} for GPT-4 and \textbf{\$1.12} for GPT-3.5. Testing with smaller step-size, like 3 plays at a time, would cost no more than \$30 for 100 games. For Claude-2.1 and Gemini-Pro, all designated tasks can be accomplished within the free usage limits.








\subsection{Applied Methods}
\label{ssec: reasoning methods}
To effectively evaluate the analytical reasoning abilities of the state-of-the-art LLMs, we employ a diversity of methodologies, including three prompting-based methods and a divide-and-conquer approach.
In the prompting-based methods, we present specifically designed prompts to language models and carefully record their responses.
Conversely, the divide-and-conquer approach involves a human-designed program to decompose the tasks into a series of elementary subtasks.
Subsequently, these subtasks are addressed individually by the language model
Finally, the responses are aggregated to formulate a comprehensive solution.

\begin{figure*}[t]
\centering
\includegraphics[width=1\linewidth]{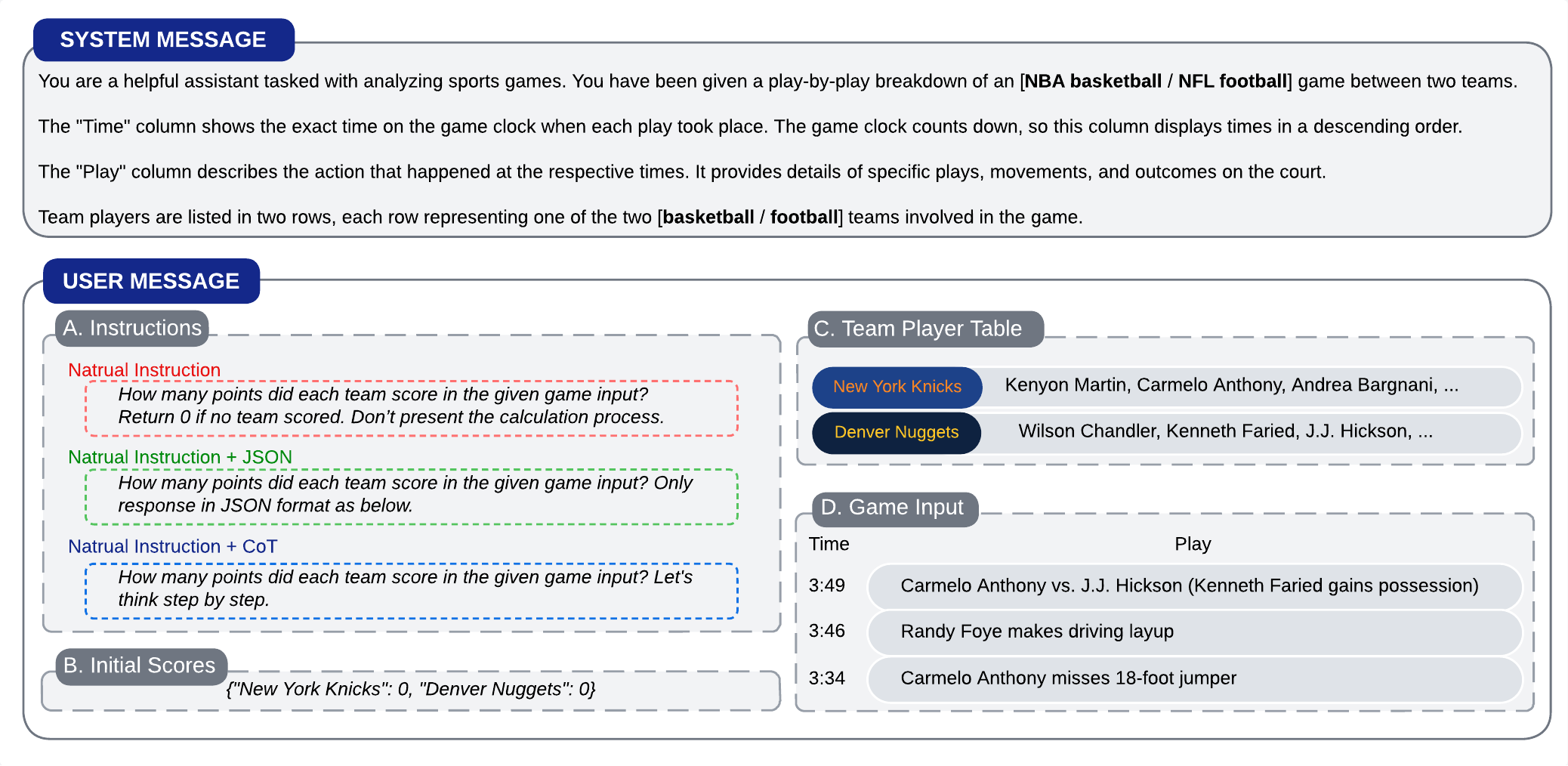}
\caption{\small We conducted comparative experiments on [A] three instructions as figure presents. After instruction, We attach sports content in order of [B] initial scores, [C] team player table and [D] game input. Aggregating the entire user message as our task instance.}
\label{fig:prompting-method}
\end{figure*}
As illustrated in Figure~\ref{fig:prompting-method}, our prompting-based methods follow a certain order of prompts. 
Initially, the prompt starts with a "\texttt{system message}" that outlines the AI system's role, contextual background, and a comprehensive description of the data formats involved.
After the "\texttt{system message}", a "\texttt{user message}" follows, which includes the user's instructions [A], the initial scores of both teams [B], details regarding player affiliations [C], and a list of the game's events [D].
Three distinct approaches are considered  to crafting user instructions:

\noindent 1) \textbf{Natural Instructions}: Utilizing natural language questions as instructions;

\noindent 2) \textbf{Natural Instructions + \textsc{JSON}}: Integrating \textsc{JSON} format with natural instructions;

\noindent 3) \textbf{Natural Instructions + CoT}: Employing a chain of thought (CoT) prompting approach to enhance the models' complex reasoning capability.

\begin{figure*}[t]
\centering
\includegraphics[width=1\linewidth]{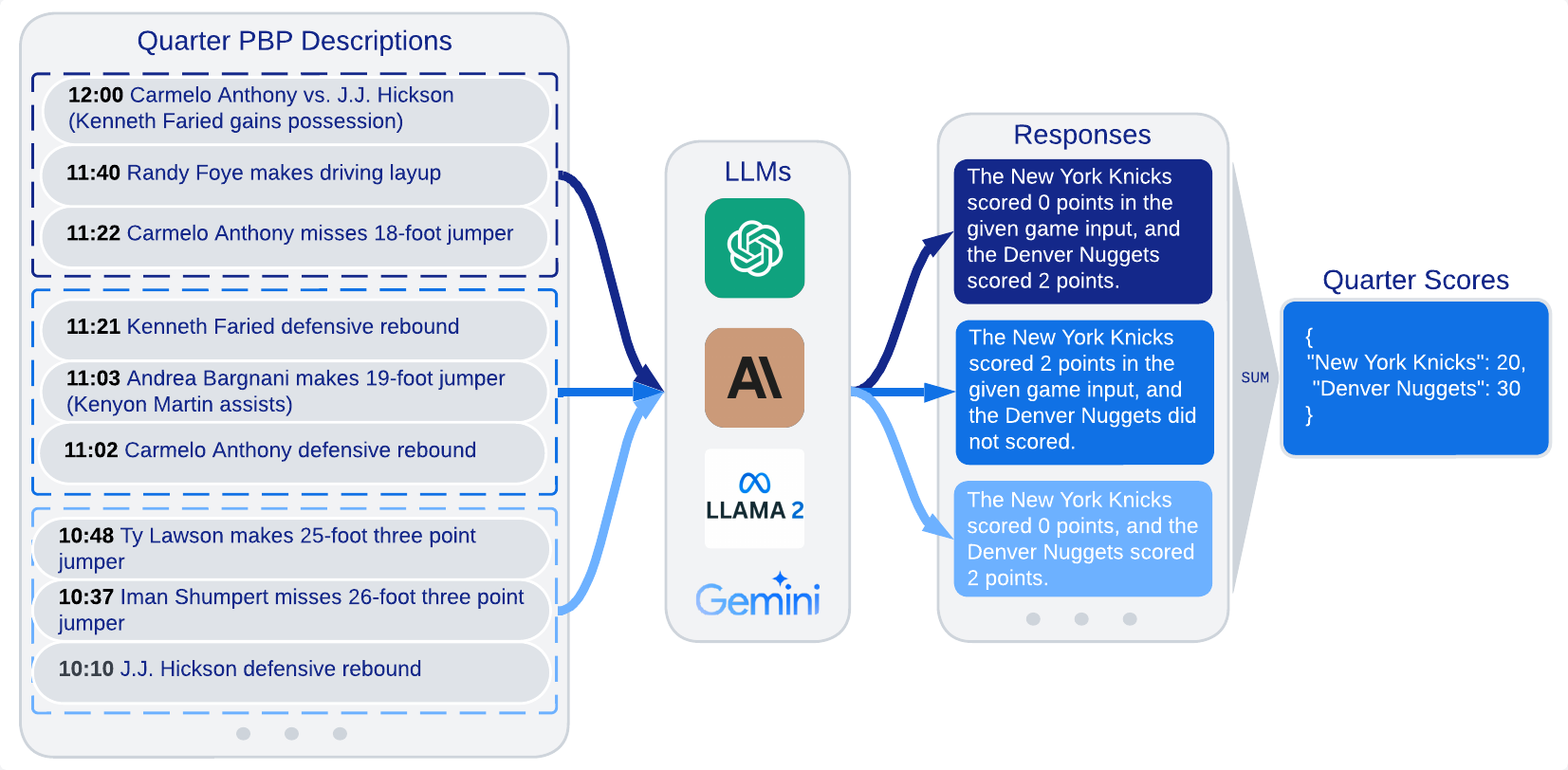}
\caption{\small A workflow depict the \textbf{divide-and-conquer} approach at step-size of three. We truncate a quarter of the play-by-play(PBP) description into segments then bind with \textbf{natural language instruction} as input to LLMs. Finally, all responses are tallied to calculate the quarter scores.}
\label{fig:method-4}
\vspace{-0.2in}
\end{figure*}
Additionally, we have developed a divide-and-conquer approach to assess reasoning capabilities.
In this approach, the sub-tasks are created by programs, while AI systems function solely as task executors.
The entire quarter is segmented into various game clips with the same number of plays $n$ in each clip except the last one.
Then, each clip is processed by LLMs, guided by the aforementioned Natural Instructions.
This process facilitates the collection of the required metrics on each clip. 
Finally, these individual metrics are aggregated to derive the overall metrics.
In Figure~\ref{fig:method-4}, an illustration is presented, demonstrating a scenario from the first quarter of an NBA game with $n=3$ plays per clip.


\section{Experiments}
\label{sec: experiments}

We conduct our experiments on 100 NBA games and 100 NFL games, focusing on the task of predicting the total scores from both teams within a given quarter. 
To evaluate the model predictions, we employ two distinct metrics:
1) Quarter-level accuracy (ACC): This metric assesses the precision of our predictions on a per-quarter basis.
A prediction is accurate if it exactly matches the total score of that quarter.
2) Quarter-level Mean Absolute Percentage Error (MAPE): This metric offers a smoothed measure of predictive accuracy by quantifying the relative deviation between the predicted and actual scores.
The formula for calculating MAPE is as follows:
\begin{equation}
    \textit{MAPE} = \frac{1}{m}\sum_{i=1}^{m}{\frac{|p_{i} - a_{i}|}{a_{i}} }
\end{equation}
Here, $m$ represents the total number of observations, $p_i$ denotes the predicted score for the $i$-th observation, and $a_i$ signifies the actual score for the $i$-th observation.
This metric is particularly valuable for understanding the magnitude of prediction errors with the actual scores, thereby providing a more detailed insight into the performance of the predictive model.

\begin{table*}[htbp]
    \centering
    \small
    \begin{tabular}{lrrrrrrrrrr}
        \toprule[1pt]
        & & \multicolumn{3}{c}{\textbf{\emph{Natural Instruction}}} & &\multicolumn{4}{c}{\textbf{\emph{Divide-and-Conquer}}}\\
        \cmidrule{3-5} \cmidrule{7-10}
        & \multicolumn{1}{c}{Models} & \multicolumn{1}{c}{NI} & \multicolumn{1}{c}{+ JSON} & \multicolumn{1}{c}{+ CoT} &  & \multicolumn{1}{c}{n=1} & \multicolumn{1}{c}{n=3} & \multicolumn{1}{c}{n=10} & \multicolumn{1}{c}{n=30}\\
        \midrule
        \multirow{5}{*}{\rotatebox{90}{\textbf{Accuracy}}} &\multicolumn{1}{c}{\small GPT-4} & 4.08 & 11.26 & \textbf{40.47} & & 20.54 & \textbf{60.14}&	38.37 &	18.31 \\
        & \multicolumn{1}{c}{GPT-3.5} & \textbf{4.45} & 3.59 & 2.37 & & 1.00 & \textbf{16.58} & 0.24 & 0.70\\
        &\multicolumn{1}{c}{\small Claude-2.1} & 3.02 & 4.33 & \textbf{16.96} & & \textbf{32.67} & 29.46 & 12.87 &	2.23 \\
        &\multicolumn{1}{c}{\small Gemini-Pro} & 2.49 & \textbf{2.83} & 2.60 & & 3.71 & \textbf{26.23} & 4.46 & 	1.48\\
        &\multicolumn{1}{c}{\small Llama-2-70b} & 3.21 & 4.34 & \textbf{4.84} & & 0.00 & \textbf{6.44} & 4.14 & 0.99 \\
        \midrule
        \multirow{5}{*}{\rotatebox{90}{\textbf{MAPE}}} &\multicolumn{1}{c}{\small GPT-4}  & 18.18 & 7.84 & \textbf{2.74} & & 30.84 & \textbf{1.70} & 2.24 & 4.86 \\
         & \multicolumn{1}{c}{\small GPT-3.5} & \textbf{14.00} & 15.99 & 19.84 & & 15.26 & \textbf{4.39} & 19.87 & 28.20 \\
        &\multicolumn{1}{c}{\small Claude-2.1} & 16.62 & 17.41 & \textbf{13.12} & & 11.74 & \textbf{5.48} & 7.65 & 14.56 \\
        &\multicolumn{1}{c}{\small Gemini-Pro} & 42.87 & 41.49 & \textbf{41.13} & & $>$100 & 14.12 & \textbf{13.69} & 35.54 \\
        &\multicolumn{1}{c}{\small Llama-2-70b} & 22.25 & \textbf{19.34} & 21.11 & & $>$100 & $>$100 & \textbf{57} & $>$100\\
        \bottomrule[1pt]
    \end{tabular}
    \caption{\small
        Total Score Results on NBA quarters.
        Both accuracy and the Mean Absolute Percentage Error (MAPE) are expressed in percentages.
        The best results achieved by each model, considering different prompting methods and various step size, are highlighted in \textbf{bold}.
        We found that LLM usually hallucinates the total scores in step by step reasoning, leading to the value of \textbf{MAPE $>$ 100}.}
    \label{tab: NBA results}
\end{table*}
Table~\ref{tab: NBA results} presents the performance results of various reasoning methods applied across different AI models in predicting NBA quarter scores.
The methods explored include three prompting-based approaches and a divide-and-conquer approach with varying step sizes ($n \in \{1, 3, 10, 30\}$).
The key observations from the table are as follows:

First, comparing the performance across all the models, GPT-4 stands out as the most effective model, with Claude-2.1 following closely behind.
The Llama-2-70b model, despite being open-sourced, performs comparably to GPT-3.5 and Gemini-Pro.
Notably, nearly all models and the combined reasoning approaches fail at the task.
Only GPT-4, when paired with the divide-and-conquer approach using a step size of $n=3$, demonstrates superior performance.
This indicates that such analytical reasoning tasks are very challenging for current LLMs.


Additionally, for prompting-based methods, the use of JSON generally enhances performance across all models.
Conversely, the Chain of Thought (CoT) approach yields mixed results; while it benefits GPT-4, Claude-2.1, and Llama-2-70b, it negatively impacts GPT-3.5 and Gemini-Pro.
This inconsistency may be attributed to the variance in training datasets among the models, suggesting that incorporating more data in the CoT style could potentially improve complex reasoning capabilities.

Subsequently, the divide-and-conquer approach, despite requiring more human efforts, tends to have better results compared to prompting only.
We observed a significant increase in Mean Absolute Percentage Error (MAPE) for Llama-2-70b and Gemini-Pro at certain step sizes.
This degradation usually indicates a significant difference between the estimated and real values.
Further investigation\footnote{See Appendix~\ref{ssec: hallucination}} revealed that these models frequently generated incorrect game scores for various single steps, which cumulatively resulted in significantly inflated aggregates.
As a result, we employ MAPE to show the variance of error predictions. And higher MAPE scores in the table indicates a greater possibility of inaccuracies.

\begin{table*}[t]
    \centering
    \small
    \begin{tabular}{lrrrrrrrrrr}
        \toprule[1pt]
        & & \multicolumn{3}{c}{\textbf{\emph{Natural Instruction}}} & &\multicolumn{4}{c}{\textbf{\emph{Divide-and-Conquer}}}\\
        \cmidrule{3-5} \cmidrule{7-10}
        & \multicolumn{1}{c}{Models} & \multicolumn{1}{c}{NI} & \multicolumn{1}{c}{+ JSON} & \multicolumn{1}{c}{+ CoT} &  & \multicolumn{1}{c}{n=1} & \multicolumn{1}{c}{n=3} & \multicolumn{1}{c}{n=10} & \multicolumn{1}{c}{n=30}\\
        \midrule
        \multirow{5}{*}{\rotatebox{90}{\textbf{Accuracy}}} &\multicolumn{1}{c}{\small GPT-4} & 55.38 & \textbf{57.38} & 55.91 &  & 95.76 & 98.25 & \textbf{98.75} & 97.00 \\
        & \multicolumn{1}{c}{GPT-3.5} & 45.38 & \textbf{50.41} & 48.00 &  & 39.50 & \textbf{81.25} & 81.25 & 79.50\\
        &\multicolumn{1}{c}{\small Claude-2.1} & 45.69 & 53.02 & \textbf{56.68} &  & \textbf{85.39} & 71.53 & 81.61 & 76.50 \\
        &\multicolumn{1}{c}{\small Gemini-Pro} & \textbf{35.23} & 35.13 & 33.02 &  & \textbf{60.50} & 51.00 & 49.23 & 50.46 \\
        &\multicolumn{1}{c}{\small Llama2-70b} & 33.54 & \textbf{35.87} & 31.53 &  & 5.29 & 9.82 & 36.75 & \textbf{52.25} \\
        \midrule
        \multirow{5}{*}{\rotatebox{90}{\textbf{MAPE}}} &\multicolumn{1}{c}{\small GPT-4}  & 24.01 & 23.92 & \textbf{25.39} && 2.31 & 1.08 & \textbf{0.66} & 1.13\\
         & \multicolumn{1}{c}{\small GPT-3.5} & 29.27 & \textbf{29.50} & 27.56 && $>$100 & 14.39 & \textbf{9.49} & 15.23 \\
        &\multicolumn{1}{c}{\small Claude-2.1} & \textbf{30.84} & 27.01 & 23.59 && \textbf{7.84} & 18.23 & 10.58 & 10.64\\
        &\multicolumn{1}{c}{\small Gemini-Pro} & 35.71 & \textbf{42.00} & 40.06 && \textbf{17.24} & 22.40 & 17.44 & 21.81 \\
        &\multicolumn{1}{c}{\small Llama2-70b} & 43.44 & 38.80 & \textbf{46.07} && $>$100 & $>$100 & $>$100 & \textbf{37.63} \\
        \bottomrule[1pt]
    \end{tabular}
    \caption{\small
    Total Score Results on NFL quarters.
    Both accuracy and the Mean Absolute Percentage Error (MAPE) are expressed in percentages.
    The best performance results in both the prompting and planning-based methods are \textbf{bolded}
    We found that Llama-2-70b and GPT-3.5 usually hallucinate a final score during the reasoning on a single step.}
    \label{tab: NFL results}
    \vspace{-0.1in}
\end{table*}
Table~\ref{tab: NFL results} compares the prediction performance of NFL quarter-level total scores across different models, which confirms the trend in model efficacy.
Remarkably, all models and reasoning strategies outperform NBA quarter prediction accuracies, which suggests that NFL score predictions may be inherently more straightforward.
Among all the prompting-based methods, Natural Instructions result in the least performance differential as compared to the other two variants.
However, as the task becomes easier, the gap between these methods is largely reduced.
In the field of divide-and-conquer methods, GPT-4  stands out with near-perfect accuracy, closely tailed by Claude-2.1 and GPT-3.5.
Despite these advancements, the issue of hallucination persists in NFL quarter predictions,  notably in the GPT-3.5 and Llama-2-70b models.


\begin{table*}[t]
    \centering
    \small
    \begin{tabular}{lr rrrrrrrrrr}
        \toprule[1pt]
        & & \multicolumn{4}{c}{\textbf{\emph{NBA}}} & & \multicolumn{4}{c}{\textbf{\emph{NFL}}}\\
        \cmidrule{3-6} \cmidrule{8-11}
        & \multicolumn{1}{c}{Models} & \multicolumn{1}{c}{n=1} & \multicolumn{1}{c}{n=3} & \multicolumn{1}{c}{n=10} &  \multicolumn{1}{c}{n=30} & &\multicolumn{1}{c}{n=1} & \multicolumn{1}{c}{n=3} & \multicolumn{1}{c}{n=10} & \multicolumn{1}{c}{n=30}\\
        \midrule
        \multirow{5}{*}{\rotatebox{90}{\textbf{Precision}}} &\multicolumn{1}{c}{\small GPT-4} & \textbf{97.85} & 97.03 & 92.28 & 66.43 & & 97.75 & 98.87 & \textbf{99.35} & 98.51 \\
        & \multicolumn{1}{c}{GPT-3.5}
        & 65.81 & \textbf{90.16} & 65.14 & 25.65 & & 53.01 & 89.79 & \textbf{92.16} & 85.63\\
        &\multicolumn{1}{c}{\small Claude-2.1}
        & \textbf{96.94} & 94.23 & 76.87 & 34.28 & & 87.27 & 86.27 & 85.73 & \textbf{88.76} \\
        &\multicolumn{1}{c}{\small Gemini-Pro}
        & 93.47 & \textbf{93.61} & 66.08 & 21.07 & & \textbf{79.66} & 71.90 & 68.54 & 58.73\\
        &\multicolumn{1}{c}{\small Llama-2-70b}
        & 10.43 & \textbf{78.18} & 53.27 & 11.43 & & 13.04 & 30.51 & 58.75 & \textbf{63.25} \\
        \midrule
        \multirow{5}{*}{\rotatebox{90}{\textbf{Recall}}} &\multicolumn{1}{c}{\small GPT-4}  & 97.22& \textbf{97.57} & 92.25& 66.41& & 98.61 & 99.11 & \textbf{99.61} & 98.22 \\
         & \multicolumn{1}{c}{\small GPT-3.5} &
        27.14& \textbf{82.23} & 62.00& 24.97& & 91.28 & \textbf{93.42} & 90.24 & 82.07 \\
        &\multicolumn{1}{c}{\small Claude-2.1} & 
        \textbf{94.31} & 93.02& 76.65& 34.17& & \textbf{93.76} & 85.35 & 89.72 & 84.29 \\
        &\multicolumn{1}{c}{\small Gemini-Pro} & 
        71.86& \textbf{93.19} & 64.34& 20.69& & \textbf{82.58} & 76.99 & 69.67 & 63.07 \\
        &\multicolumn{1}{c}{\small Llama-2-70b} & 
        45.38& \textbf{80.25} & 53.45& 10.21& & 79.42 & 77.36 & \textbf{82.05} & 66.81\\
        \bottomrule[1pt]
    \end{tabular}
    \caption{\small
       To further explore what factors, in divide-and-conquer method, could influence the accuracy on entire quarter. We compute \textbf{precision} and \textbf{recall} on scoring plays. We notice that precision results are more correlated to the accuracy evaluation on quarter level. 
        }
    \label{tab: step level results}
    \vspace{-0.2in}
\end{table*}
Moreover, in both Table~\ref{tab: NBA results} and Table~\ref{tab: NFL results}, interestingly, we found that a step size of one does not necessarily achieve the best performance when employing a divide-and-conquer approach.
Actually, the optimal step sizes for different models and tasks are different.
This phenomenon can be attributed to the fact that a smaller step size usually requires a larger amount of total steps, leading to an accumulation of errors in individual steps.
Therefore, choosing an appropriate step size is essential for optimizing quarter-level accuracy, while balancing single-step prediction accuracy and the total number of steps required for these tasks.
For a more fine-grained understanding of results obtained by the divide-and-conquer approach with different step sizes, we further calculate the precision and recall for scoring plays on step-level predictions.
As presented in Table~\ref{tab: step level results}, surprisingly, using a step size of one may not necessarily yield the optimal performance even on the step level.
This can be attributed to the system messages and instructions potentially overshadowing the critical play-by-play data when the step size is minimal.
We usually observe that models have higher precision scores than recall scores on scoring plays, except for the LLama-2-70b model.
This may indicate that such models are more likely to mistakenly predict clips without any scoring play.
\section{Analysis}
\label{sec: analysis}

The experiments described in Section~\ref{sec: experiments} highlight the complex nature of sports reasoning tasks.
In pursuit of a deeper understanding of the complexities associated with these problems,
we undertake a series of analyses.
Through a methodical approach of hypothesis formulation and subsequent verification, we aim to dissect the multifaceted factors contributing to task difficulties.


\subsection{Length Factor}
\label{ssec: length factor}
Initial observations suggest an increase in task difficulties correlated with length, as evidenced by the decline in performance of the divide-and-conquer methods with increasing step sizes, detailed in Table~\ref{tab: NBA results}.
To further investigate this phenomenon, we expanded our experimental analysis to include a broader range of lengths for NBA games, specifically 2-minute intervals, quarters, half-games, and full games.
For a more focused analysis, the study was limited to examining prompting-based methods.

\begin{wrapfigure}{r}{0.48\textwidth}
    \begin{center}
        \includegraphics[width=0.43\textwidth]{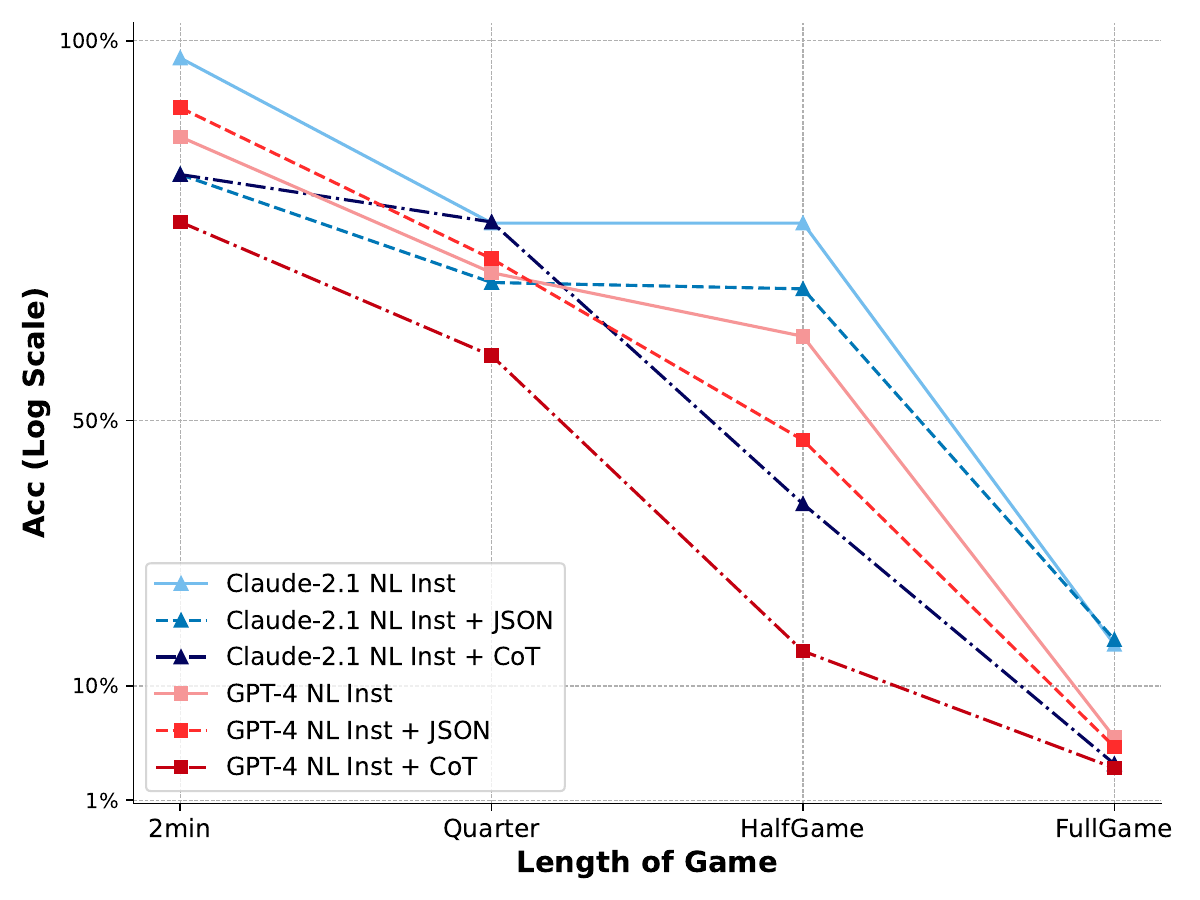}
    \end{center}
    \caption{\small Length Analysis on NBA data}
    \label{fig:length-results}
\end{wrapfigure}

The performance trends of three such methods across different length scales are plotted in Figure~\ref{fig:length-results}.
To mitigate the impact of model limitations, this analysis exclusively considers the top-2 models, namely GPT-4 and Claude-2.1.
Based on the analysis of the plot, it is observed that all curves exhibit a downward trend as the context length increases.
This deviation could potentially be attributed to the prevalence of whole game scores in news reports, suggesting that these outliers may result from background knowledge acquired during the model's training phase.
Consequently, it is reasonable to infer that within the confines of the same task, the complexity of the task escalates in correlation with an increase in context length.

\subsection{Information Density Factor}
\label{ssec: density factor}
Another interesting observation is the discrepancy in the performance between the NBA and the NFL games.
Upon examination of Table~\ref{tab: data statistics}, it is evident that the intrinsic characteristics of the two sports differ significantly, with basketball generally featuring a higher frequency of scoring plays in comparison to American football.
Consequently, we propose a hypothesis that the complexity of the task may be influenced by the information density in the provided context.
Here, we define information density as the ratio of scoring plays to the total number of plays within a given game clip.
\begin{figure*}[htbp]
\centering
\includegraphics[width=\linewidth]{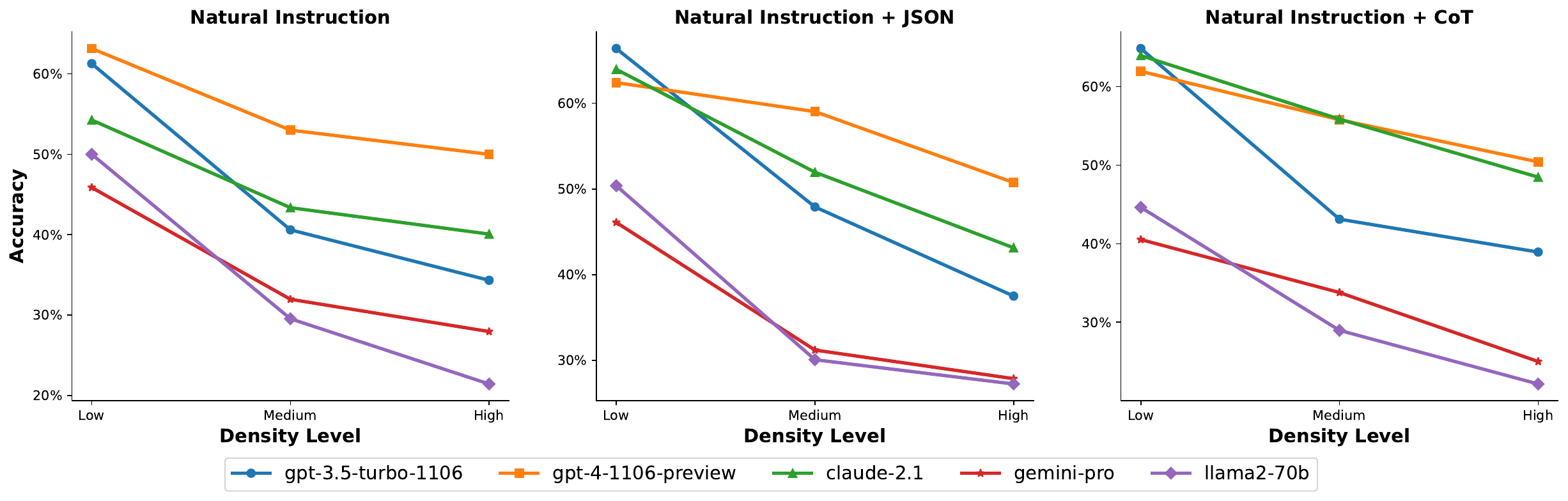}
\caption{\small Density analysis on NFL data}
\label{fig:density-results}
\end{figure*}
Figure~\ref{fig:density-results} illustrates the tendency in model prediction accuracy across distinct levels of information density within NFL quarters.
The data points are classified into three tiers: low density, medium density, and high density, corresponding respectively to the bottom one-third, middle one-third, and top one-third of the information density values.
Across all models and reasoning methods, a consistent trend emerges indicating a decline in model performance with increasing information density levels.
This observation allows us to infer that the complexity of the reasoning task escalates in direct proportion to the level of information density.

\subsection{Related Information Factor}
\label{ssec: related information factor}
It is important to decide whether the information provided to the model for analytical reasoning should be restricted to the necessary details or include related but non-essential information. 
When calculating the total number of scores from both teams, the players and the teams are the essential entities that are involved in the reasoning process.
However, their names are not required for the reasoning steps.
You can replace their names with any other synonymous identifiers without affecting the outcome or the logical progression of the reasoning.
Thus, we propose to replace the names of these entities with identifiers such as "\texttt{Player-1}", "\texttt{Player-2}", and "\texttt{Team-1}, \texttt{Team-2}" to evaluate the impact of such modifications on the model's predictions.
\begin{table*}[htbp]
    \centering
    \small
    \begin{tabular}{lrrrrr}
    \toprule[1pt]
    \multicolumn{1}{c}{Information Provided} & \multicolumn{1}{c}{GPT-4} & \multicolumn{1}{c}{GPT-3.5} & \multicolumn{1}{c}{Claude-2.1} & \multicolumn{1}{c}{Gemini-Pro} & \multicolumn{1}{c}{Llama-2-70b} \\
    \midrule
    Whole & 40.47~~ & 2.37~~ & 16.96~~ & 2.60~~ & 4.84~~ \\
    \quad - Team Name & 40.84\textcolor{green}{$\uparrow$} & 2.10\textcolor{red}{$\downarrow$} & 14.73\textcolor{red}{$\downarrow$} & 3.22\textcolor{green}{$\uparrow$} & 3.06\textcolor{red}{$\downarrow$} \\
    \quad - Player Name & 32.30\textcolor{red}{$\downarrow$} & 2.35\textcolor{red}{$\downarrow$} & 18.56\textcolor{green}{$\uparrow$} & 2.85\textcolor{green}{$\uparrow$} & 2.58\textcolor{red}{$\downarrow$} \\
    \quad - Both & 28.58\textcolor{red}{$\downarrow$} & 2.84\textcolor{green}{$\uparrow$} & 15.84\textcolor{red}{$\downarrow$} & 1.60\textcolor{red}{$\downarrow$} & 3.79\textcolor{red}{$\downarrow$}\\
    \bottomrule[1pt]
    \end{tabular}
    \caption{\small Assessing the impacts of related sports information on reasoning performance. All values in the table are presented in percentages(\%). The evaluation was conducted on quarter-level scores inference utilizing chain of thought(CoT) prompting.}
    \label{tab: related information}
\end{table*}
Table~\ref{tab: related information} showcases the performance of the Chain of Thought (CoT) prompting approach in calculating total scores across different models, where part of the entity names have been replaced, specifically within the context of NBA quarters.
After analyzing the data in the table, we have found that in most cases, the performance of the model decreases if some parts of the entity names are missing.
However, we have also discovered a few cases where the model's performance improves.
It indicates that while the model should be trained more robustly to handle missing information, adding relative but non-essential information generally helps to enhance performance.
This may be because such information acts as pre-conditions for the next token predictions, leveraging the model's internal knowledge, therefore narrowing down the option of the prediction.

\section{Conclusion}
\label{sec: conclusion}

In this paper, we investigate the analytical reasoning capabilities of LLMs by analyzing play-by-play data from NBA and NFL games, focusing on the models' ability to compute the total points scored by each team.
Despite advancements in LLMs, such as OpenAI's GPT series and others like Claude, Gemini, and Llama, these models generally struggle with sports analytical reasoning tasks.
This study reveals significant disparities in performance among different models, with GPT-4 showing somewhat better results in specific scenarios, especially when using techniques like Chain of Thought prompting.

Further, our research explores the impact of different methodologies on model performance, including the Chain of Thought prompting and a divide-and-conquer strategy that segments data into smaller, more manageable parts.
These methods aim to enhance the models' analytical capabilities, but the study reveals that, despite these efforts, LLMs still struggle with accurately processing and analyzing sports statistics, indicating a gap in their ability to handle complex reasoning tasks effectively.

Additionally, the extensive analysis showcases the factors that affect the task complexity.
These include the length, information density, and presence of relative information.
It suggests, that to develop a better system, we should focus more on enhancing the models' data-processing capabilities.
To improve the understanding and use of complex information, it may be necessary to collect more diverse and intricate training data, including Chain of Thought style data.
Additionally, models should be developed to be more robust to both relevant and irrelevant information.
Finally, building the model to work effectively with precise tools such as code and databases is essential.


\bibliography{colm2024_conference}
\bibliographystyle{colm2024_conference}

\appendix
\section{Appendix}
\label{sec: appendix}

\subsection{Hallucinated Responses}
\label{ssec: hallucination}

\begin{figure}[h]
\centering
\includegraphics[width=0.8\linewidth,height=5cm]{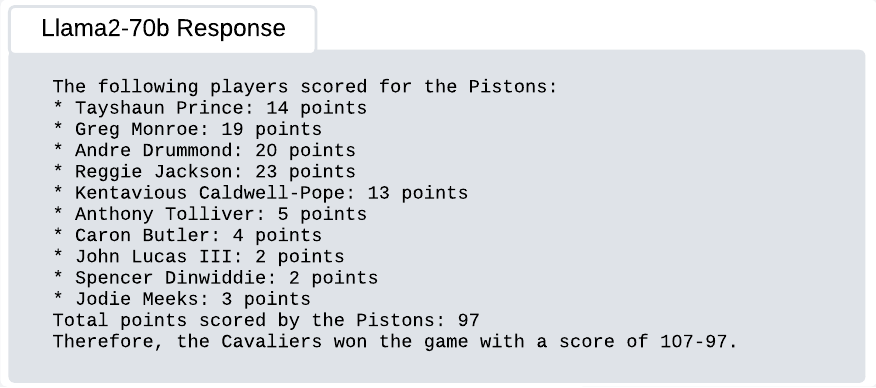}
\caption{\small Hallucinated response when task contains only one sport description (step\_size=1)}
\label{fig:hall-llama2}
\end{figure}

We addressed one particular type of hallucination is overestimation of scores as shown in \ref{fig:hall-llama2}. 

We analyze model outputs from NBA divide-and-conquer experiments. Once the model’s prediction exceeds the threshold, defined by maximum points gained in ground truth at a certain step size, we identify it as hallucinations. We found that Llama-2-70b exhibited a significant error rate of \textbf{46.84\%} at step size = 1 which directly results in oddly high MAPE as we presented in Table \ref{tab: NBA results}. The close-source models responsed with fewer than 1\% of hallucinations.

\begin{figure}[h]
\centering
\includegraphics[width=1\linewidth]{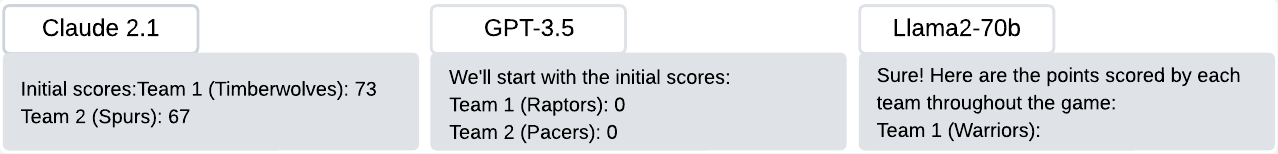}
\caption{\small LLMs predict a team name when player and team identities were masked.}
\label{fig:hall-team}
\end{figure}
When we assessing the impacts of related sports information, an unusual response caught our attention as presented in Figure \ref{fig:hall-team}. In this task, all teams and player should be masked out. But LLMs includes GPT-3.5, Claude-2.1 and Llama2-70b predict a team name in their responses and reasoning based on it. In analysis results from Table \ref{tab: related information}, those models present a close or even better reasoning performance over natural game input. Meanwhile, other models, GPT-4 and Gemini-Pro, which didn't predict related sports information shows a great performance degradation.

\subsection{Instruction Following Verification}

\begin{table}[htbp]
    \centering
    \begin{tabular}{p{0.9\textwidth}}
    \toprule
    Error response example - From Llama2-70b\\
    \midrule
    Sure! Here are the points scored by each team at the end of the 2nd quarter: \\
    \\
    Brooklyn Nets: 47 points \\
    Memphis Grizzlies: 55 points \\
    \\
    The Grizzlies have a 8-point lead over the Nets at the end of the 2nd quarter \\
    \bottomrule
    
    \end{tabular}
    \caption{\small Responses not includes reasoning steps given CoT prompt.}
    \label{tab: error_cot}
\vspace{-0.1in}
\end{table}

\begin{figure}[h]
\centering
\includegraphics[width=1\linewidth, height=9cm]{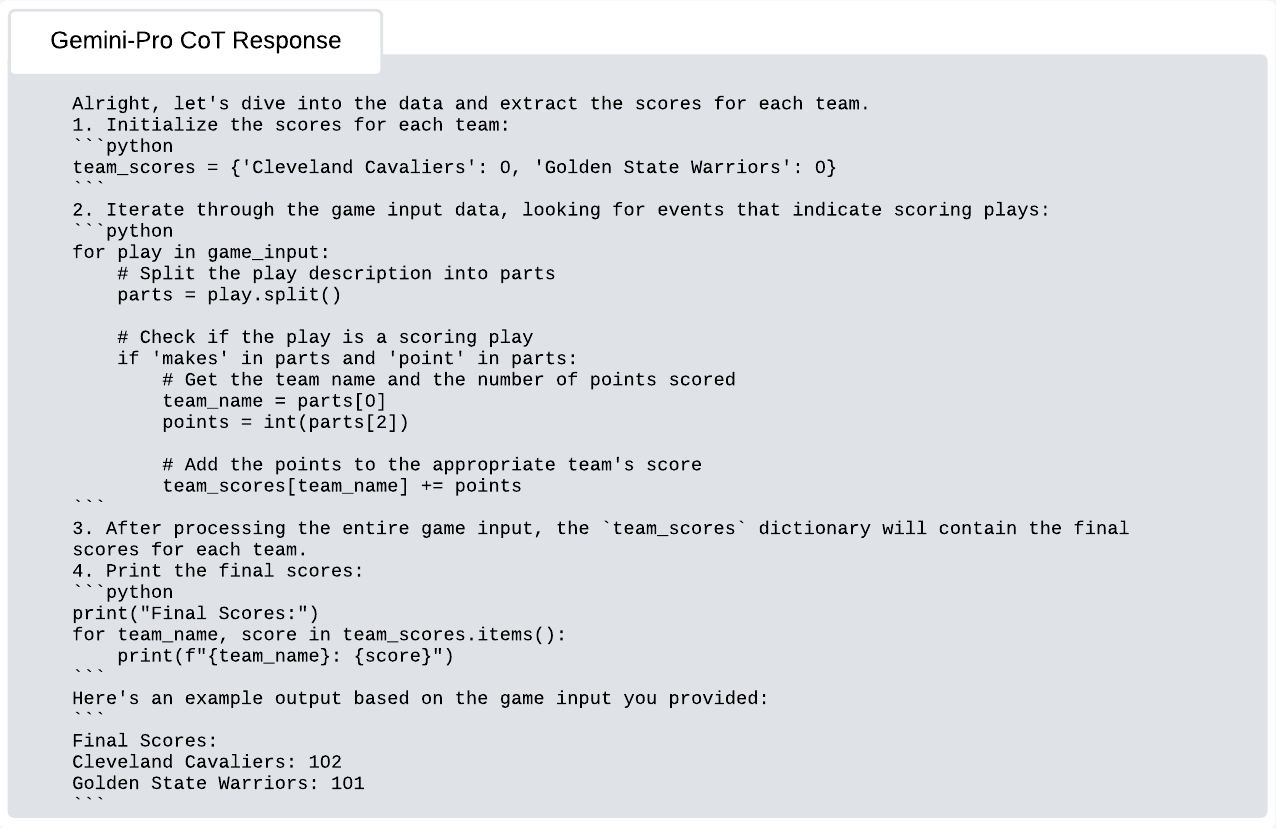}
\caption{\small Example from HalfGame reasoning task with CoT instruction. Gemini-Pro is the only model respond with Python script. We regard this as error presentation of middle steps since it does not showing the math reasoning procedure.}
\label{fig:gemini-cot-exp}
\end{figure}

\begin{figure*}[h]
\centering
\includegraphics[width=1\linewidth, height=7cm]{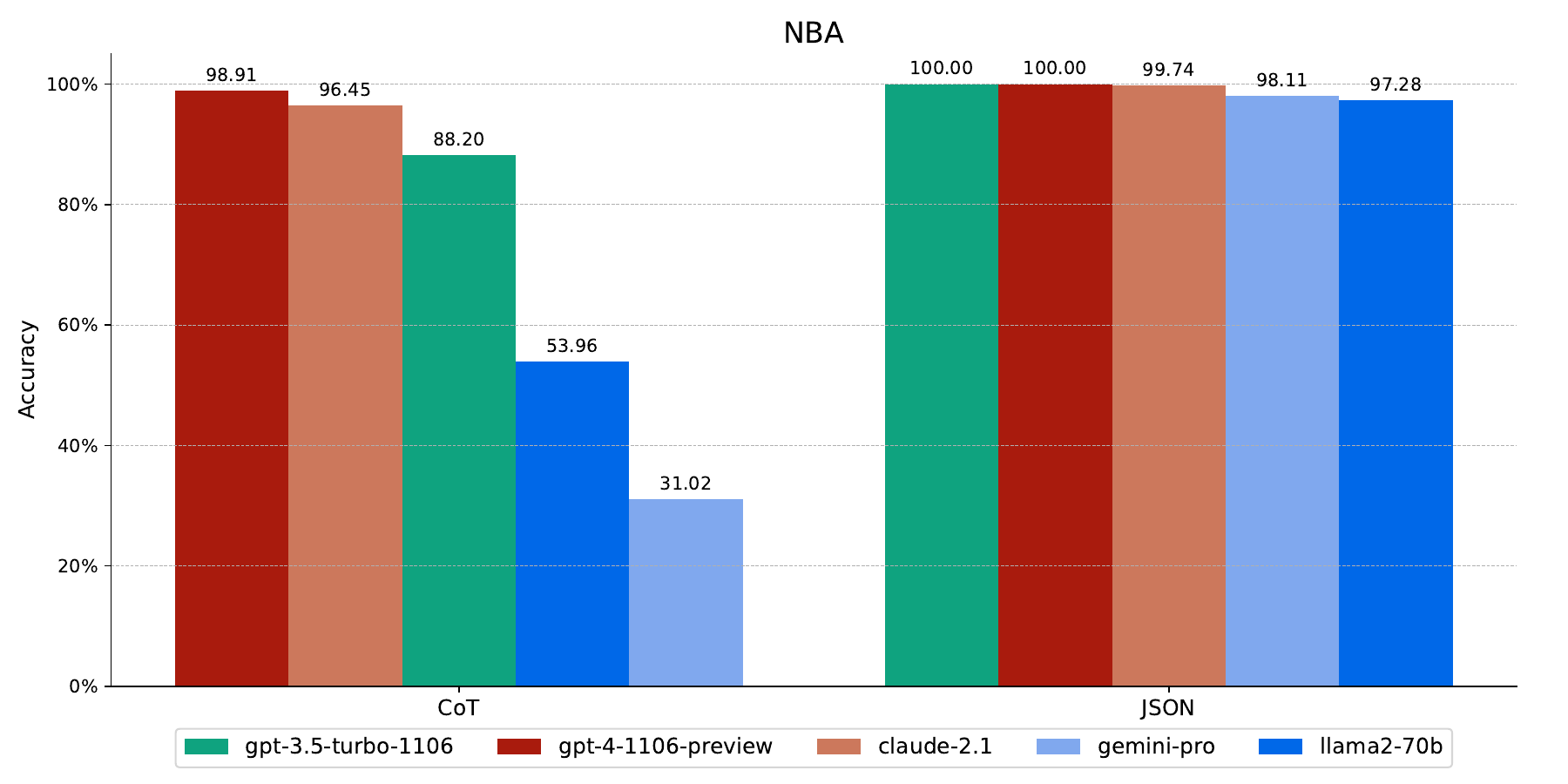}
\caption{\small Instruction following accuracy bar chart for NBA tasks. We calculated number of non-zero digits in the response to assess whether LLM perform step-by-step reasoning process. For NBA tasks, we set 10 as threshold and we analyze the model's instruction following ability through trends.}
\label{fig:nba-inst-following}
\end{figure*}

\begin{figure*}[h]
\centering
\includegraphics[width=1\linewidth, height=8cm]{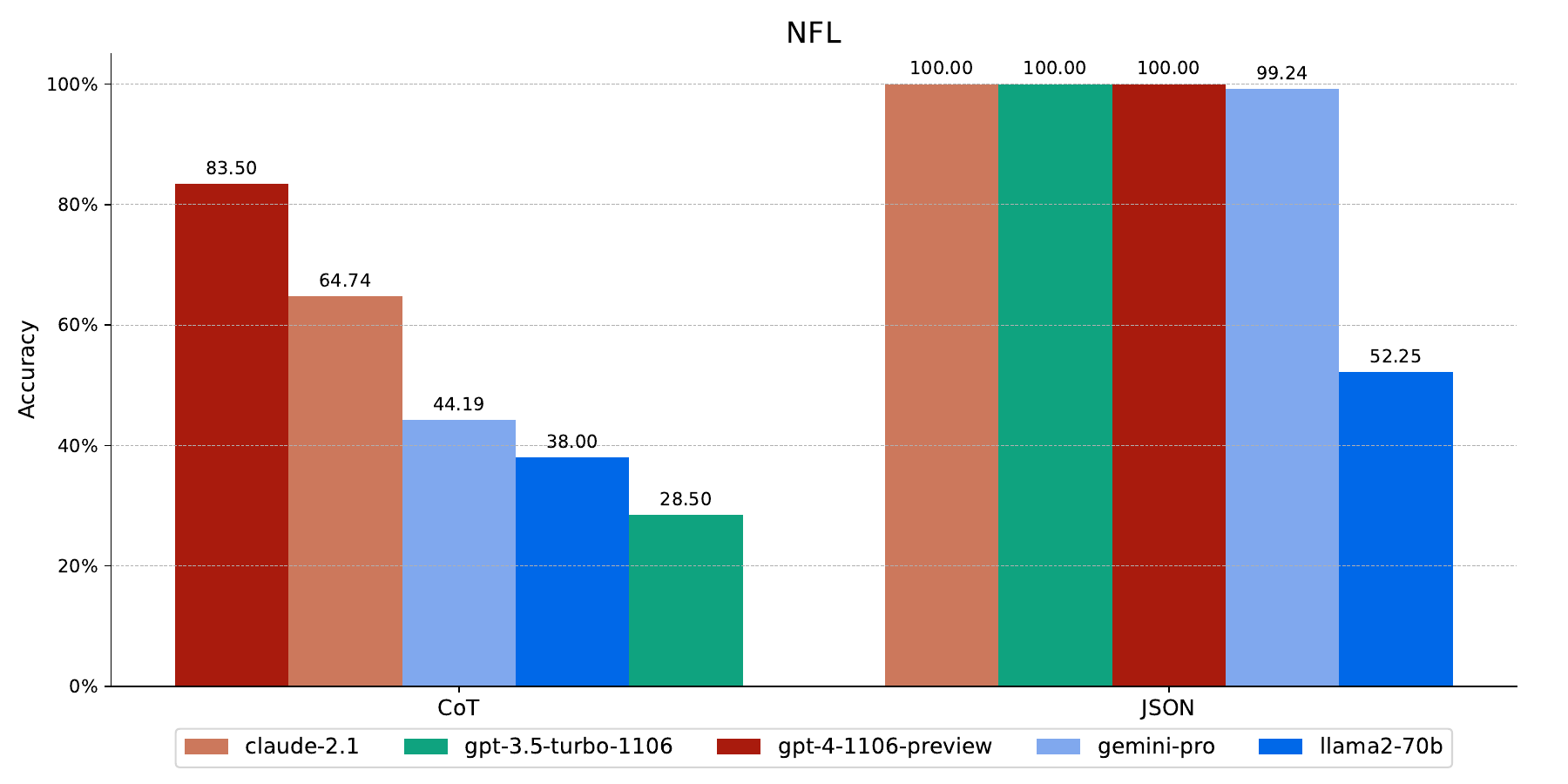}
\caption{\small Instruction following accuracy bar chart for NBA tasks. We applied same justification method for CoT accuracy. For NFL tasks, we set 6 as threshold because less scoring moves included in game input.}
\label{fig:nfl-inst-following}
\end{figure*}

The content in Table \ref{tab: error_cot} is an example of a common error response occurred when we use CoT method. We present another type of error response in Figure \ref{fig:gemini-cot-exp}. Finally, We post the overall accuracy comparison across all tasks whichever requires JSON formatting or CoT instruction following capability in Figure \ref{fig:nba-inst-following} and Figure \ref{fig:nfl-inst-following}.

\end{document}